\documentclass[conference,twocolumn]{style/IEEEtran}
\usepackage[utf8]{inputenc}
\usepackage[T1]{fontenc}
\usepackage[english]{babel}
\usepackage{amsmath,amssymb,bbm}
\usepackage{graphicx}
\usepackage{float}
\usepackage{cuted}
\usepackage{placeins}
\usepackage{afterpage}
\usepackage{xfrac}
\usepackage{tikz}
\usepackage{xcolor}
\usepackage{natbib}
\usepackage{microtype}
\usepackage{dblfloatfix}
\usepackage{balance}
\usepackage{caption}
\definecolor{mydarkblue}{rgb}{0,0.08,0.65}
\usepackage[colorlinks=true,linkcolor=mydarkblue,citecolor=mydarkblue,filecolor=mydarkblue,urlcolor=mydarkblue]{hyperref}
\usepackage[nameinlink,noabbrev]{cleveref}
\usepackage{titletoc}

\setcitestyle{round}
\makeatletter
\def\@IEEEsectpunct{\enspace}
\renewcommand{\paragraph}{\@startsection{paragraph}{4}{\z@}%
  {1.5ex plus 0.5ex minus 0.2ex}{-1em}{\normalfont\normalsize\itshape}}
\makeatother

\newcommand{\zyphraaffiliation}{\textsuperscript{1}\,Zyphra Research}
\newcommand{\equalcontribution}{{\mdseries\footnotesize *\,Equal contribution}}
\newsavebox{\mainpointbox}
\newenvironment{mainpoint}{%
  \par\addvspace{\medskipamount}%
  \begingroup
  \setlength{\fboxrule}{0.5pt}%
  \setlength{\fboxsep}{6pt}%
  \begin{lrbox}{\mainpointbox}%
  \begin{minipage}{\dimexpr\linewidth-2\fboxsep-2\fboxrule\relax}%
  \normalfont\normalsize\ignorespaces
}{%
  \end{minipage}%
  \end{lrbox}%
  \noindent\fbox{\usebox{\mainpointbox}}\par
  \endgroup
  \addvspace{\medskipamount}%
}
\crefname{figure}{Figure}{Figures}
\crefname{table}{Table}{Tables}
\crefname{section}{Section}{Sections}

\title{How Local Mixing Encodes Relative Position in Global NoPE Attention}
\author{%
Cutter Dawes\textsuperscript{1,*}, Nick Alonso\textsuperscript{1,*}, Tom Figliolia\textsuperscript{1}, Beren Millidge\textsuperscript{1}\\[0.35em]
\zyphraaffiliation \quad \equalcontribution%
}

\begin{document}
\maketitle
\setcounter{page}{1}
\begin{abstract}\normalfont\mdseries
The attention operation is naively position invariant. However, positional information is fundamental to natural language, and therefore a variety of explicit position encodings have been developed in transformer-based models, such as rotary position encoding (RoPE). Although explicit position encodings have long been assumed to be required, recent methods that interleave local mixing layers, such as sliding window attention (SWA) and gated linear attention, while \textit{not} encoding position (NoPE) in global attention layers has recently been shown to be successful at scale. How and why this approach works is not well-understood. In this paper, we develop an explanation of how hybrid models of this sort can implicitly encode position at global NoPE layers. Supported by both theoretical and empirical evidence, our central argument is that SWA and gated linear attention induce a recency bias in the residual stream that propagates to, and is selected by, the global attention logits. Moreover, in contrast to the implicit position encodings found in models with only global NoPE attention, in which positional information arises solely from the causal mask, the recency bias in hybrid models can be maintained across long sequences. In addition to deepening our understanding of how hybrid models encode position, these findings may provide insights for how to encode position in a way that can extrapolate to longer sequence lengths indefinitely.
\end{abstract}

\section{Introduction}

Distinguishing tokens by their position is a vital component of language modeling \citep{dufter2022position}. Since self-attention by itself is permutation invariant, attention layers in large language models (LLMs) typically utilize position encodings (PEs), which explicitly represent either the absolute or relative position of queries and keys. However, commonly used PEs such as rotary positional encoding \citep[RoPE;][]{su2024roformer} incur computational overhead and make length extension and extrapolation difficult. Removing explicit PE (a method called NoPE) has met with mixed success: NoPE does not match the in-domain performance of the same architecture with explicit PE, although it can provide some improvements in length extrapolation whereas the performance of RoPE and other explicit position embeddings tend to collapse quickly beyond their training sequence length \citep{haviv2022transformer,kazemnejad2023impact}.

Separately, to mitigate the computational and memory costs of attention across long sequences, so-called \textit{hybrid} architectures that interleave either sliding window attention (SWA) or gated linear layers with global attention layers are increasingly used at scale. Interestingly, as hybrid models have become more widespread, there has been a shift in the global attention layers from using RoPE \citep{gemmateam2024gemma2} to p-RoPE \citep{qwen2025qwen3next} to, most recently, NoPE. Notably, Kimi K3 interleaved global NoPE with a gated linear variant called KDA \citep{kimit2026k3}.

However, there are remaining questions concerning how such models utilize and represent position in their NoPE attention layers, an issue that is critical to their long context performance. To our knowledge, only one prior paper has studied how hybrid models represent position in NoPE layers, and their main finding is that they do not represent \textit{absolute} position \citep{puvvada2025swan} at all. However, this result leaves open the question of whether such models may learn to encode \textit{relative position} in their NoPE layers.

In this paper, we present novel theoretical and empirical evidence that hybrid architectures can and do develop relative PEs at their NoPE layers. Unlike other \textit{explicit} relative PEs such as RoPE, hybrid architectures \textit{implicitly} encode relative position via the interactions between local or linear layers and subsequent NoPE layers. Specifically, the SWA or gated linear layers introduce a recency bias in the residual stream that propagates to, and is selected by, the global NoPE attention logits and then accentuated through depth. We support this account in two complementary ways:
\begin{enumerate}
    \item[(i)] Theoretically, we show mathematically how SWA approximates a moving-average convolution in expectation, inducing a recency bias in its output that transfers into the residual stream, propagates through intervening layers, and is selected by the global NoPE attention logits.
    \item[(ii)] Empirically, we demonstrate the existence of the predicted recency bias in the residual stream and NoPE attention logits across depth. We show this bias is present at initialization and strengthens throughout training. We validate this in hybrid models interleaving global NoPE attention with SWA (with RoPE and NoPE) and KDA (Appendix~\ref{app:kda})
\end{enumerate}

\begin{figure*}[!htbp]
    \centering
    \includegraphics[width=0.90\textwidth]{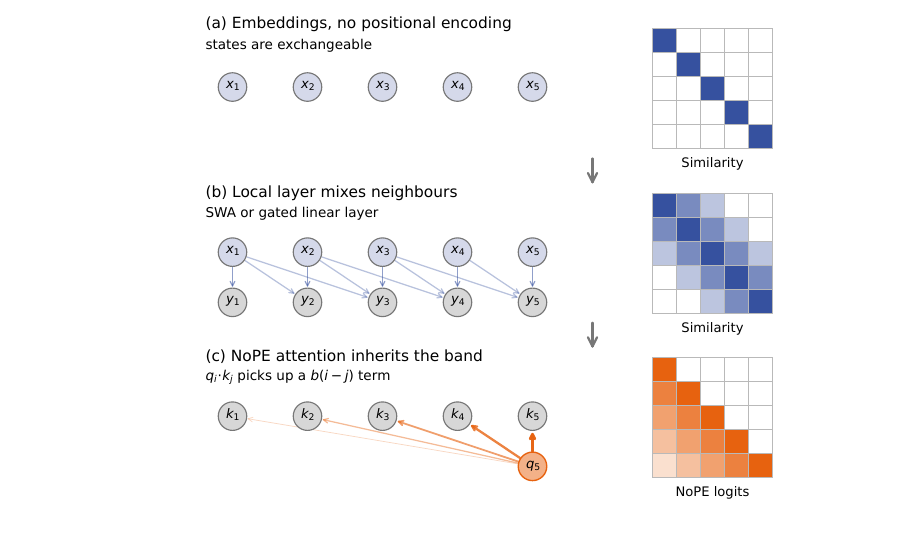}
    \caption{\textit{Proposed mechanism for implicit relative position encoding.} (a) Tokens begin as initially uncorrelated. (b) Local mixing correlates nearby residual states. (c) Learned query and key projections can read out this lag-dependent structure as recency-biased global NoPE logits. The shading intensity denotes correlation or logit strength, while line width denotes mixing or attention weight.}
    \label{fig:hybrid-mechanism}
\end{figure*}

An overview of our hypothesized mechanism is provided in Figure~\ref{fig:hybrid-mechanism}. In the following sections, we first note that existing relative PEs generally induce a recency bias (Section~\ref{sec:background}), then we introduce our core argument in three steps (Section~\ref{sec:thesis}). First, SWA produces a residual stream recency bias (Section~\ref{sec:step1}); second, this recency bias propagates through the intervening modules such as normalization and multi-layer perceptrons (MLPs; Section~\ref{sec:step2}); and third, this recency bias propagates to and is selected by the global NoPE attention logits (Section~\ref{sec:step3}). 

\section{Background: Recency Bias as Relative Position Encoding} \label{sec:background}

Relative PEs, as opposed to absolute PEs, are the industry standard in LLM architectures \citep{shaw2018self,dufter2022position,su2024roformer}. These PEs alter attention logits, directly or indirectly, in a way that depends on the relative distance between query and key. There are a variety of methods for encoding relative positions in attention, but the majority share the following feature:
\begin{mainpoint}
    \textit{Common and state-of-the-art relative position encoding methods encode relative position by creating a recency bias in attention logits}. That is, given a query $q_i$ at position $i$, and a key $k_j$ at $j$, relative position encodings bias the attention logits downward as the relative distance $i-j$ increases.
\end{mainpoint}
Relative PEs encoding a recency bias is not strictly necessary \citep{chen2025hope}. In principle, a relative PE applied to attention can induce any kind of bias in attention logits, as long as this bias depends directly on relative position. However, using a recency bias to encode relative position appears to work well in practice, because: (i) it provides a particularly simple positional signal for the model to learn and use (i.e., in which distance is represented directly by a smooth, monotonic change in the attention strength); and (ii) it fits the statistics of natural sequential data \citep[i.e., the tokens most important for predicting next tokens tend to be the most recent ones;][]{tan2025stick,wu2025emergence,kim2026layernorm}.

\paragraph{Survey of relative PEs.}
There are generally two ways to create a relative PE in attention logits: additively or multiplicatively \citep{zhang2026grape}. Additive encodings directly add a bias, $b_{i-j}$, to attention logits; i.e., $s = q_i^{\top}k_j + b_{i-j}$, where $i$ and $j$ are the position of query and key and $i-j$ the relative position. Examples include Alibi and Fire \citep{press2022train,li2024fire}, which add a fixed, data-independent bias 
that decays linearly with relative distance, as well as the more recent Fox PE \citep{lin2025forgetting} that uses a data-dependent bias. All of these explicitly induce a recency bias in attention logits, such that keys farther from the query have a larger negative bias added to their attention logits. Multiplicative PEs include a matrix, $R_{i-j}$, in the query-key multiplication; i.e., $s = q_i^{\top}R_{i-j}k_j$, where $R_{i-j}$ is unique to relative position $i-j$. RoPE, arguably the best-known relative PE, rotates pairs of channels using rotation matrices in both the queries and keys. It has been observed that learned attention heads with RoPE either focus on very low-frequency channels with little resulting influence on logit decay, or they focus on high-frequency channels in such a way that they create a strong recency bias \citep{barbero2025round}. Other multiplicative PEs, like the recent Wall attention \citep{yang2025path}, create a similar kind of recency bias in a way that is data-dependent \citep{yang2025path,zhang2026grape}.

\paragraph{PE in NoPE.}
Among previous work studying the position encoding resulting from NoPE, two papers are particularly relevant to this work. First, \citet{puvvada2025swan} studied implicit position encodings in models with global NoPE attention, including hybrid architectures that interleave global NoPE with SWA. In particular, they trained probes to predict absolute position encoding in the residual stream embeddings, finding that the probes could predict absolute position in global NoPE architectures but not in hybrid architectures (see \citealp{haviv2022transformer,chi2023latent} for more evidence of absolute position encodings in global NoPE architectures). This provides evidence that hybrid NoPE architectures, unlike their global NoPE counterparts, do not encode absolute position, but leaves open the possibility that hybrid architectures could encode relative positions. 

Second, \citet{zuo2025position} showed that architectures with global NoPE attention at every layer could learn a recency bias-based PE in their residual stream, created by the asymmetric token mixing effects of attention with a causal mask. However, their analysis and empirical results only demonstrate a recency bias effect on very short sequences (approximately 30 tokens), leaving open the question of how this bias may change with sequence length. Additionally, they do not explain the conditions under which attention can make use of this recency bias in the residual stream. In this work, we apply a similar analysis to hybrid architectures with SWA layers and compare to global NoPE models. Importantly, we show that SWA has a recency bias effect on the residual stream independent of total sequence length, whereas for initialized global NoPE residuals the recency bias loses resolution even at moderate sequence lengths. Further, unlike previous works, we explain the conditions required of attention parameters to make use of this recency bias in the residual stream, and we scale our empirical results to models with 350M+ parameters trained on natural language data.

\section{Thesis: Hybrid Architectures Implicitly Encode Relative Position at Global NoPE Layers} \label{sec:thesis}

We develop and support the following thesis about hybrid architectures with global NoPE attention:
\begin{mainpoint}
    \textit{Local layers (e.g., SWA) output a sequence of vectors which, in expectation, have a recency bias (i.e., smaller cosine similarity with greater relative distance). These outputs then create a recency bias in the post-attention residual stream that is persistent across sequence lengths. This residual stream bias can (and does after training) get transferred through the intervening layers and to attention logits at NoPE layers, thereby creating an} implicit \textit{relative position encoding in NoPE layers}.
\end{mainpoint}

We provide two lines of evidence that present complementary views of this effect. The theoretical analysis isolates the structural effect of one type of local layer (SWA) and outlines how the resulting recency bias can propagate through intervening modules and into the global NoPE logits. However, our analysis does not fully account for the variety of computation the model is being trained for, particularly the effects of content-dependent attention and MLPs. Our empirical analysis then verifies the existence of the residual stream and logit-based recency biases in multiple models, at varying stages of training, on both random and natural language data (for random-token controls, see Appendix~\ref{app:random-controls}). We also show how these effects generalize to hybrid architectures that use KDA (Appendix~\ref{app:kda}) as the local layer instead of SWA. Below, we first provide theoretical and then empirical preliminaries common to each step in our proposed mechanism.

\subsection{Theory} \label{sec:thesis-theory}

First, we formally introduce several notions of similarity in the residual stream, and what we mean by a recency bias. Let $X=[x_1,\dots,x_T] \in \mathbb{R}^{D \times T}$ be the residual stream at some layer. The most simple and intuitive measurement of similarity is cosine similarity; i.e., $c(d) = \mathbb{E}[\sfrac{x_i^\top x_{i-d}}{\|x_i\| \|x_{i-d}\|}]$. However, cosine similarity captures just one slice of the full similarity structure; in particular, it does not account for cross-channel correlations that may be picked up by the global attention projections. Therefore, we also introduce the lagged cross-moment matrix, $G_X(d)=\mathbb{E}[x_i x_{i-d}^\top]$, where the $x_i$ are unit-normalized.

To obtain scalar measurements of similarity, one can consider various slices of this matrix; i.e., the Frobenius inner product $\langle G_X(d), M \rangle_F = \operatorname{tr}(G_X(d)^\top M)$ with some matrix $M$. Different slices of the lagged cross-moment correspond precisely to our similarity measures of interest: cosine similarity is the isotropic component, and the NoPE logits capture alignment with the composed query-key projection. That is,
\[ \begin{aligned}
    \text{Cosine:}\quad c(d) &= \langle G_X(d), I\rangle_F, \\
    \text{NoPE logits:}\quad \ell(d) &= \langle G_Z(d), M\rangle_F.
\end{aligned} \]
Here, $I$ is the identity matrix, $Z$ is the actual post-normalization input to a NoPE head at its (unnormalized) model scale, and $M=W_Q^\top W_K/\sqrt{D_H}$ with head dimension $D_H$. Given one of the above scalar similarity profiles $f(d)$, we compute the recency gap as $f(1)-f(L)$, where $L$ is the far-distance reference lag (4096 in our experiments); then, $f(L)$ is the similarity floor, or the underlying correlation structure in the sequence present at long distances.


\subsection{Empirics} \label{sec:thesis-empirics}

For the empirical analysis, we study hybrid architectures at both the 120M and 350M scales, varying the SWA window size $w\in\{64,128,256,512,1024,2048,4096\}$ and comparing to a global NoPE baseline (i.e., no SWA interleave) at each scale. Unless otherwise noted, the SWA layers use RoPE and are interleaved with global NoPE at a $1{:}1$ ratio. The main text shows results for the 350M models, and the $w=128$ variant for single-model analyses. We train the models on the Prolong dataset \citep{gao2025prolong}. For evaluations periodically throughout training, we use TextbookChapters \citep{chevalier2024language}.

Using the measures defined in Section~\ref{sec:thesis-theory}, we track residual stream cosine similarity and global NoPE attention logits across training, depth, and SWA window size. For more details on the experimental setup, including training setup, statistics sampling, and head aggregation, see Appendix \ref{app:methods}.
\section{Step 1: SWA Produces a Recency Bias in the Residual Stream} \label{sec:step1}

\subsection{Theory} \label{sec:step1-theory}

Unlike full attention, SWA restricts its attention operation to a local window of keys and values. Here, we show that this local operation gives the attention output a distance-dependent cosine similarity profile. Furthermore, because its scale is fixed by the window size rather than the sequence length (in contrast to full attention), this effect does not dilute as the total sequence length grows. 

For a single attention head, define $q_i = W_Q x_i$, $k_i = W_K x_i$, $v_i = W_V x_i$. For a given $q_i$ and $k_{i-d}$, the attention weight at distance $d$ is
\[
    \alpha_i(d)
    =
    \frac{\exp\!\left(q_i^\top k_{i-d}/\sqrt{D_H}\right)}
    {\sum_{r=0}^{w-1}\exp\!\left(q_i^\top k_{i-r}/\sqrt{D_H}\right)},
    \qquad 0\leq d<w,
\]
where $w$ is the sliding window size and $D_H$ the head dimension. Writing $o_i = W_O W_V x_i$, the SWA output is 
\[
    y_i = W_O \sum_{d=0}^{w-1} \alpha_i(d) v_{i-d} = \sum_{d=0}^{w-1} \alpha_i(d) o_{i-d}.
\]
Hence, SWA acts akin to a data-dependent convolution. 

\paragraph{Intuition.}
To illustrate why such a local filter may produce a recency bias in its output, consider a simplified scenario that closely reflects SWA in an LLM at initialization, in which the projected values $o_i$ are zero-mean, of similar magnitude, and uncorrelated across positions, while the attention weights are roughly uniform. Then we may approximate the SWA output as a moving average,
\[
    y_i \approx \frac{1}{w} \sum_{r=0}^{w-1} o_{i-r}.
\]
In this simplified scenario, the cosine similarity of two outputs, say $y_i = \tfrac{1}{w} (o_{i-w+1} + \dots + o_i)$ and $y_{i-d} = \tfrac{1}{w} (o_{i-d-w+1} + \dots + o_{i-d})$, depends on the fraction of projected values shared by their sums. The closer they are in position, the larger this shared fraction; e.g., for a window size of 64, $y_i$ shares 63 terms with $y_{i-1}$, 62 with $y_{i-2}$, and so on. Thus, the expected output cosine similarity decreases with relative distance and approaches zero once the windows no longer overlap.

\paragraph{Length scaling and comparison with global NoPE.}
As noted in Section~\ref{sec:background}, \citet{zuo2025position} did a similar analysis, but for global NoPE architectures; i.e., they showed that a global NoPE attention layer could induce an adjacency pattern in its outputs. However, they considered only tiny sequences ($\sim30$ tokens) and did not explore whether this effect scaled with sequence length. Here, the uniform-attention toy case helps explain why global NoPE cannot sustain a well-resolved recency bias even out to 1000-token sequences. Specifically, the output of a global NoPE layer will be 
\[
    y_i \approx \frac{1}{i+1} \sum_{t=0}^{i} o_{t}.
\]
Unlike SWA, the model averages over the entire preceding sequence. Therefore, if $i=1000$, then $y_i$ and $y_{i-10}$ share $99\%$ of their terms, so the uniform-attention cosine changes only slightly over fixed lags. Together, these calculations predict a persistent window-scale recency profile for SWA, but not a comparably resolved fixed-lag profile for global NoPE.

To make the contrast with SWA more concrete, under the same initialization-like assumptions as above, the output's cosine similarity with distance is:
\[
\begin{aligned}
    c_\text{SWA}(d) &\approx \big[ 1 - \frac{d}{w} \big]_+, \\
    c_\text{global}(d) &\approx \sqrt{(i-d+1)/(i+1)} \propto 1 - O(d/i).
\end{aligned}
\]
Hence, at initialization, the strength of SWA's recency bias decays with the window size $w$, with a smaller window resulting in a stronger, steeper recency bias; conversely, global attention's positional signal decays with the sequence length $i$, and therefore dilutes for longer sequences. That is, SWA's locality scale is length-independent by construction, whereas global attention's is not.

\stepcounter{figure}
\afterpage{%
\noindent\begin{minipage}{\columnwidth}
\centering
\includegraphics[width=\columnwidth]{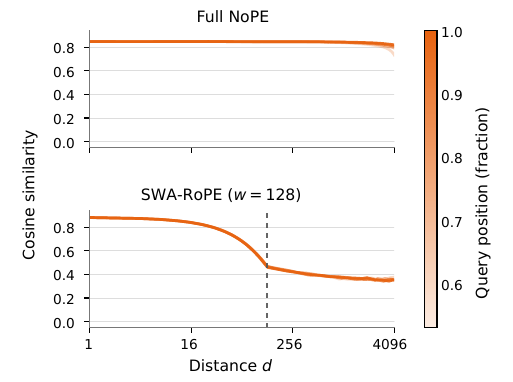}
\addtocounter{figure}{-3}
\captionof{figure}{\textit{Attention outputs at initialization.} Raw cosine of the isolated first attention-branch outputs for global NoPE and SWA ($w=128$; marked by dashed line), across four fixed query-position bins. The recency bias in the SWA output is stronger and remains constant with increased sequence length, whereas the recency bias of full attention weakens significantly at longer sequences.}
\label{fig:initial-attention-outputs}
\addtocounter{figure}{2}
\end{minipage}\par\vspace{\textfloatsep}%
}

\begin{figure*}[!b]
\centering
\includegraphics[width=0.95\textwidth]{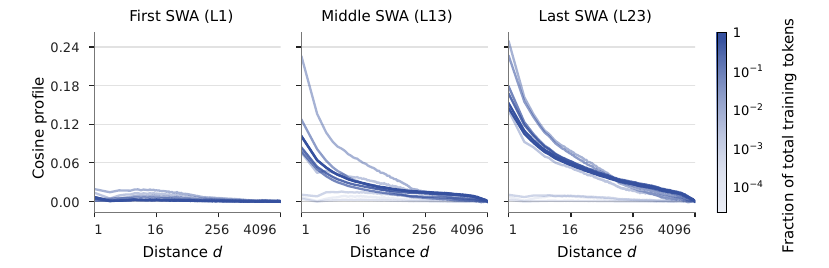}
\caption{\textit{Residual cosine during training.} Post-attention cosine profiles at the first, middle, and last SWA layers ($w=128$), with color denoting the fraction of training completed on a logarithmic scale. This demonstrates that residual stream recency bias increases with training and depth.}
\label{fig:residual-training}
\end{figure*}

\begin{figure*}[!t]
\centering
\includegraphics[width=0.95\textwidth]{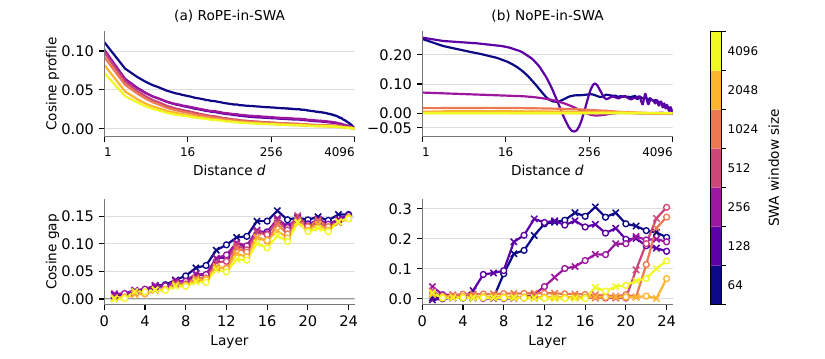}
\caption{\textit{Window dependence of residual cosine.} (a) RoPE-in-SWA and (b) NoPE-in-SWA, with the  top panels showing the cosine gap $c(d)-c(4096)$ at the middle SWA layer (L13), and the bottom panels showing the cosine gap $c(1)-c(4096)$ across depth (crosses mark SWA layers and circles global NoPE layers). With or without RoPE, smaller windows result in stronger recency biases across depth.}
\label{fig:residual-window-gaps}
\end{figure*}

\paragraph{General mean-attention case.}
We now describe how SWA affects the similarity structure in its output after training, where the attention profile is no longer uniform and the incoming residual stream no longer zero-mean Gaussian. Again, we approximate attention by its mean profile $a_d = \mathbb{E}[\alpha_i(d)]$ (estimated across some number of relevant sequences). Suppose that $X$ has some incoming lagged cross-moment structure $G_X(d)$; let $A = W_O W_V$ be the composed projected-value matrix and $g_a(k) = \sum_s a_s a_{s+k}$ the mean attention autocorrelation. In this setting, the SWA output is $y_i \approx A \sum_{d=0}^{w-1} a_d x_{i-d}$, and its lagged cross-moment is
\[
    G_Y(d) \approx A \sum_k g_a(k) G_X(d-k) A^\top = A (g_a \ast G_X)(d) A^\top,
\]
if output norms are concentrated (see Appendix~\ref{app:step1-theory} for the full derivation). Thus, the similarity structure in the output is the incoming structure mixed across distances by the mean attention autocorrelation, and then mixed across channels by the composed attention matrices. The structural effect of SWA is still present, as $g_a(k) > 0$ if $|k|<w$, and 0 otherwise. So, the window bounds the scale of the recency bias, but its effect can be amplified or dampened by the attention profile within the window. Finally, the extent to which this mixing promotes or dilutes a recency bias in the output with respect to each similarity metric depends on its alignment to the relevant projections.

\subsection{Empirics} \label{sec:step1-empirics}

\paragraph{Initialization and length scaling.}
Empirically, we first validate the argument that SWA's effect is structurally different from that of global NoPE, and that only the former produces a well-resolved recency bias at long sequence lengths. Figure~\ref{fig:initial-attention-outputs} compares the isolated first attention-branch outputs before residual addition at initialization. With increasing sequence lengths (probed via query position along an 8k-token sequence), global NoPE activations become so highly correlated that positions cannot be resolved for most distances; in contrast, SWA retains a strong distance-dependent profile across the window while flattening beyond, regardless of sequence length.

\paragraph{Emergence with training and depth.}
Though the analysis above corroborates the hypothesis that SWA produces a recency bias at initialization, it is possible that this effect dissipates during training as the model focuses on learning content-dependent interactions. However, this is not the case -- in fact, the recency bias strengthens during training. Figure~\ref{fig:residual-training} visualizes the recency profile during training at the first, middle, and last SWA layers. Across depth, the trained profiles show strengthening and subsequent attenuation relative to their peak, while remaining stronger than at initialization. Note that this effect stabilized early in training (the colorbar is logarithmic in the fraction of that scale's total training tokens). Furthermore, the recency bias strengthens significantly with increased depth. Further analysis as well as layer-by-layer profiles are shown in Appendix~\ref{app:swa-depth-windows}.

\paragraph{Window-size dependence.}
The theoretical analysis in Section~\ref{sec:step1-theory} suggests that smaller SWA windows will produce a narrower and steeper recency bias in the outgoing residual stream. The distance profiles (top) and layer-wise recency gaps (bottom) in Figure~\ref{fig:residual-window-gaps}a verify this trend; smaller windows lead to stronger recency biases across depth (for full profile comparisons, see Figure~\ref{fig:residual-depth-windows-350m}b in Appendix~\ref{app:step1}). Further, to mitigate the confound of the strong logit-based recency bias induced by RoPE within the window (which as we discussed in Section~\ref{sec:step1-theory} has an important contribution to the recency bias), we also verify this for models with NoPE-in-SWA. Figure~\ref{fig:residual-window-gaps}b shows the distance profiles and layer-wise recency gaps across window sizes for NoPE-in-SWA; in fact, the effect of window size is more pronounced, with smaller windows producing significantly stronger recency biases (for further NoPE-in-SWA results, please see Figure~\ref{fig:120m-robustness-loss-composite}f in Appendix \ref{app:controls}).

Given the pronounced effect of SWA window size on recency bias strength, a further question one might ask is how window size affects model performance. We find that smaller windows achieve lower loss than their larger-windowed counterparts (see Figure~\ref{fig:swa-loss-windows-350m} here and Figure~\ref{fig:120m-robustness-loss-composite}g--h in the appendix, which report the late-training validation-loss trajectories; note that larger NoPE-in-SWA windows cause loss collapse). Although preliminary, this is promising evidence that the recency bias strength is important to such hybrid models' success. In fact, the benefit of strengthening the recency bias effect by decreasing window size may outweigh the cost of the associated reduction in training compute so that the performance of the model improves on shorter window sizes. Also note that, since these experiments did not match compute across window sizes, the advantage for small windows is even stronger than shown here.

\begin{figure}[!t]
\centering
\includegraphics[width=\columnwidth]{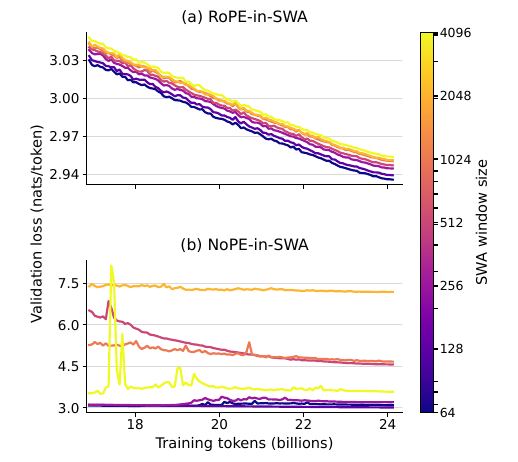}
\caption{\textit{Validation loss across window sizes.} (a) RoPE-in-SWA and (b) NoPE-in-SWA. Mean 350M DCLM token cross-entropy (nats) over the final 30\% of training. With or without RoPE, smaller windows result in lower loss.}
\label{fig:swa-loss-windows-350m}
\end{figure}

\section{Step 2: Preservation by Intervening Transformations} \label{sec:intervening-transformations} \label{sec:step2}

\begin{figure*}[!b]
\centering
\includegraphics[width=0.9\textwidth]{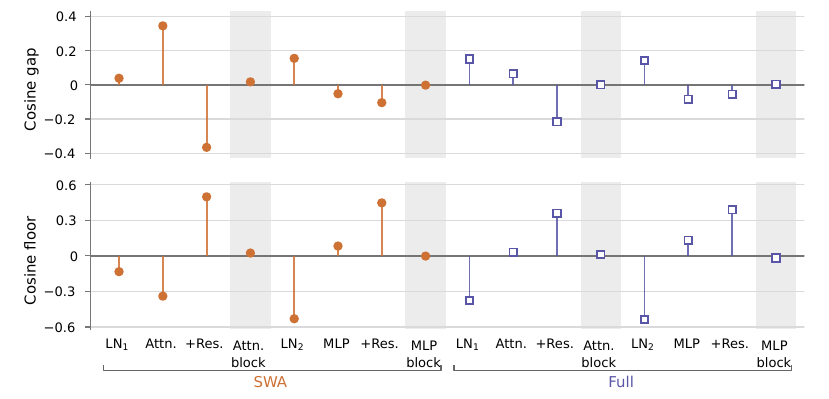}
\caption{\textit{Module changes in cosine gap and floor.} After-minus-before changes in cosine gap $c(1)-c(4096)$ (top) and cosine floor $c(4096)$ (bottom), averaged across SWA and full-attention blocks at $w=128$. Residual-addition steps compare with the isolated branch; shaded block summaries compare the complete update with its incoming residual. Note that the effect of each module on the similarity structure is largely as predicted by our theory.}
\label{fig:module-cosine-gap-floor}
\end{figure*}

\begin{figure*}[!t]
\centering
\includegraphics[width=0.95\textwidth]{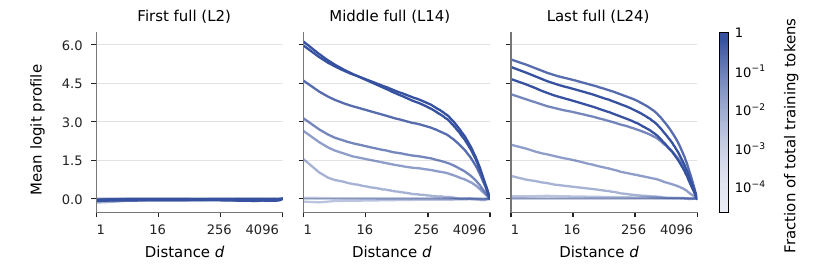}
\caption{\textit{Global NoPE logits during training.} Head-mean profiles at the first, middle, and last global layers ($w=128$), with color denoting the fraction of training completed on a logarithmic scale. This shows that, through training, the global NoPE heads learn to select the residual stream recency bias and thereby develop a corresponding recency bias in their logits.}
\label{fig:logit-training}
\end{figure*}

\subsection{Theory} \label{sec:step2-theory}

\paragraph{Addition of the residual stream.}

Suppose there is some similarity structure $G_Y(d)$ in the SWA output $Y$, and $G_X(d)$ in its input $X$. Denote the following residual stream by $Z = X + Y$; we wish to understand its similarity structure $G_Z(d)$ as a function of its components. Expanding the outer-product in $G_Z(d) = \mathbb{E}[z_i z_{i-d}^\top]$, we obtain
\[
    G_Z(d) = G_X(d) + \mathbb{E}[x_i y_{i-d}^\top] + \mathbb{E}[y_i x_{i-d}^\top] + G_Y(d)
\]
For uncorrelated white inputs, the first and second terms are 0, and the third term becomes $a_d A \Sigma$, and so we obtain $G_Z(d) = a_d A \Sigma + g_a(d) A \Sigma A^\top$. In either case, the similarity structure $G_Y(d)$ is carried into the residual stream, and the resulting similarity structure depends on its interactions with the existing similarity structure $G_X(d)$.

\paragraph{Effect of normalization.}
The actual normalization operation, whether this is the (numerically stable, scaled by $\sqrt D$) L2 normalization for RMSNorm or first subtracting the channel mean for LayerNorm, has little effect on similarity structure. RMSNorm in fact exactly preserves the unit-normalized cross-moment structure; and, though LayerNorm additionally projects prior to L2 normalization, prior work has found that they behave very similarly in practice \citep{zhang2019root, gupta2026geometric}.\footnote{In particular, \citet{gupta2026geometric} found that the residual streams of transformers with LayerNorm are approximately orthogonal to the one-vector in the models studied, which they show to be the normal vector of the projection prior to normalization.}

In contrast, the gain (and offset for LayerNorm; though we omit that here for simplicity) of the normalization module can alter the similarity structure, including its floor and gap. Here, let $X$ denote the output of the normalization stage. Decompose $X$ into its variation around a position-independent mean $\bar x = \mathbb{E}[x_i]$, so that each $x_i = \bar x + \epsilon_i$ where $\mathbb{E}[\epsilon_i] = 0$. For per-channel gains $D_\gamma = \text{diag}(\gamma)$ and output $Y = D_\gamma X$, we have
\[
    \mathbb{E}[y_i y_{i-d}^\top] = \underbrace{D_\gamma \bar x \bar x^\top D_\gamma}_{\text{shared-mean contribution}} + \underbrace{D_\gamma \mathbb{E}[\epsilon_i \epsilon_{i-d}^\top] D_\gamma}_{\text{fluctuation cross-moment}}.
\]
Write $M=D_\gamma\bar x\bar x^\top D_\gamma$ for the shared-mean contribution and $e_M=\operatorname{tr}(M)=\|D_\gamma\bar x\|^2$ for its energy, and $C(d)=D_\gamma \mathbb{E}[\epsilon_i \epsilon_{i-d}^\top] D_\gamma$ for the fluctuation cross-moment and $e_C=\operatorname{tr}(C(0))=\mathbb{E}[\|D_\gamma\epsilon_i\|^2]$ for its energy. Assuming approximately stationary second moments and concentrated norms, the unit-normalized similarity gap and far-distance floor are
\[
    G_Y(d)-G_Y(L) \approx \frac{C(d)-C(L)}{e_M+e_C}, \qquad G_Y(L) \approx \frac{M+C(L)}{e_M+e_C}.
\]
How the normalization module's gain affects the similarity gap and floor (and in turn, the cosine similarity, or another projection or magnitude of the cross-moment) depends on its effect on the shared-mean contribution and fluctuation cross-moments and their energies. The gain's effect on each of these terms depends in turn on how the channel-pair entries are reweighted by $\gamma_c\gamma_k$, with the energies depending on the diagonal weights $\gamma_c^2$. In practice, the broad pattern appears to be that the normalization's gain reduces the shared-mean energy relative to the fluctuation energy; when the fluctuation profile $C(d)/e_C$ is approximately preserved, this can reduce the cosine similarity floor while amplifying the similarity gap.

\paragraph{Effect of the MLP.}
For normalized inputs, a wide MLP at random initialization admits a simple kernel description of its effect on cosine similarity \citep{daniely2016toward}; i.e., at initialization an MLP preserves the ordering of the recency-bias profile.

Suppose the MLP branch is $y=W_2f(W_1x)$, with wide random Gaussian $W_1,W_2$, and consider two inputs $x,x'$ with cosine similarity $p$. For each hidden unit, the corresponding preactivations $U=w^\top x$ and $V=w^\top x'$ are jointly Gaussian with correlation $p$. We therefore reduce the problem to asking how the nonlinearity $f$ transforms this correlation, defining the resulting kernel $k_f(p)$; the subsequent projection $W_2$ approximately preserves this kernel at large width. Expanding $f$ in the Gaussian Hermite basis gives
\[
k_f(p)=\frac{\sum_n a_n^2p^n}{\sum_n a_n^2}.
\]
The $n=0$ term contributes a constant similarity floor, the $n=1$ term transmits similarity linearly, and higher-order terms nonlinearly reshape the profile. Since the coefficients enter as $a_n^2$, $k_f$ is nondecreasing for $p\ge0$. Thus, at random initialization, a wide MLP preserves the ordering of a nonnegative recency-bias profile, although it may change its floor, magnitude, and shape.

This initialization argument does not guarantee preservation by trained MLPs. Depending on the projections and nonlinearities learned in training, the MLP can suppress, amplify, or reorient the incoming similarity structure. We therefore assess these effects empirically in Section~\ref{sec:step2-empirics}, for the similarity gap and floor each.

\subsection{Empirics} \label{sec:step2-empirics}

We observed in Section~\ref{sec:step2-theory} that residual stream addition, normalization, and MLPs each have a complex effect on the full cross-moment similarity structure, including both its gap and floor. To empirically characterize these effects, Figure~\ref{fig:module-cosine-gap-floor} measures the difference in the cosine gap and floor, before and after each module (averaged across each of its appearances in depth; the grey block summaries show the cumulative effect across each block). Proceeding through each module appearance from left to right, we generally observe the following trends: (i) normalization, including its learned affine transformation, reduces the cosine floor and increases the gap; (ii) SWA provides the strongest boost to the gap (and interestingly, reduces the floor); (iii) the residual addition reduces the gap and increases the floor; (iv) the MLP reduces the gap and increases the floor; and (v) global NoPE attention also increases the gap, though much less so than SWA. The cumulative effect across each block is much smaller, with full-attention and MLP blocks largely preserving the incoming gap.

\section{Step 3: Global NoPE Inherits the Residual Stream Recency Bias} \label{sec:step3}

\subsection{Theory} 

Suppose the recency bias has been preserved until the input to global NoPE attention; the final step is its successful readout into the global NoPE logits. Let $x_i$ denote the post-normalization input to global NoPE's query and key projections. For a fixed attention head, let $M = W_Q^{\top}W_K/\sqrt{D_H}$ and recall $G_X(d)=\mathbb{E}[x_i x_{i-d}^\top]$. The expected logit profile is
\[
    \ell(d) = \mathbb{E}[x_i^\top M x_{i-d}] = \langle G_X(d),M\rangle_F.
\]
Thus, the logit recency gap $\ell(1)-\ell(L)$ is positive precisely when $M$ aligns positively with $G_X(1)-G_X(L)$; its magnitude depends on the extent of that alignment. Notably, in contrast to cosine, this similarity profile can depend on the skew component of the full cross-moment structure. Depending on the slice orientation, the projections can preserve, distort, or reverse the residual-similarity ordering (see Appendix~\ref{app:step3} for further discussion).

Note also that, for independent zero-mean query and key initializations, $\mathbb{E}[M]=0$; therefore, the recency bias does not immediately propagate to the full attention logits in expectation over initialization. However, training can align $W_Q$ and $W_K$, allowing their logits to inherit the residual stream recency bias, which predicts that the effect should become visible in NoPE logits as training induces favorable alignment.

\subsection{Empirics} \label{sec:step3-empirics}

\paragraph{Emergence with training.}
To empirically verify this, Figure~\ref{fig:logit-training} tracks the global NoPE attention logits during training. Indeed, at initialization the distance dependence is weak; but, as training progresses, a recency-biased profile emerges across depth, consistent with learning the projection alignment described above. Similarly to the development of the residual stream recency bias following SWA, the recency bias in the full attention logits strengthens and stabilizes very early in training (the colorbar is logarithmic in the fraction of training completed). But, whereas the residual stream recency bias increases consistently with depth, the logit recency bias quickly strengthens then stabilizes (layer-by-layer profiles are shown in Appendix~\ref{app:logit-depth-windows}).

\begin{figure}[!h]
\centering
\includegraphics[width=\columnwidth]{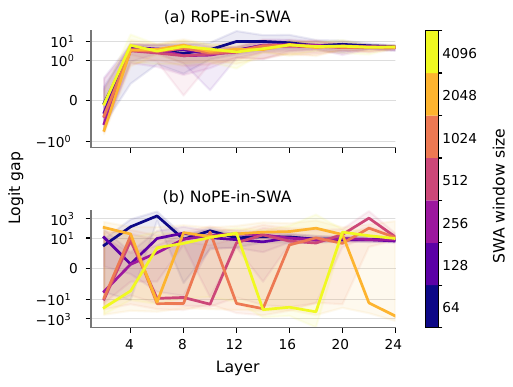}
\caption{\textit{Global NoPE logit gaps.} RoPE-in-SWA (top) and NoPE-in-SWA (bottom), showing the head-mean gap $\bar\ell(1)-\bar\ell(4096)$ (shading spans the full headwise range and vertical axes are symmetric-log). Though the window dependence is less clear in RoPE-in-SWA, it appears the logit recency bias collapses for large windows with NoPE-in-SWA.}
\label{fig:logit-gap-windows}
\end{figure}

\paragraph{Window-size dependence.}
We found in Section~\ref{sec:step1} that smaller windows induce a stronger residual stream recency bias. To investigate the extent to which this trend propagates into the global NoPE logits, Figure~\ref{fig:logit-gap-windows} summarizes the layer and head distribution of the recency gap. In comparison to the residual stream, a few distinct patterns emerge. For both RoPE-in-SWA and NoPE-in-SWA, the logit gap remains fairly stable across depth after initially strengthening (as seen above). For RoPE-in-SWA (Figure~\ref{fig:logit-gap-windows}a), there is also little variation by window size; however, for NoPE-in-SWA (Figure~\ref{fig:logit-gap-windows}b), the logit gaps for larger windows swing drastically positive and negative (note that the recency gap axis is symmetric-log scaled to account for the wider variation). Further analysis as well as layer-by-layer profiles are shown in Appendix~\ref{app:step3-empirics}.

These patterns suggest that the model seeks to learn a consistent logit recency bias across depth. However, if window size is too large (and therefore, the residual stream recency bias too weak) then the model is unable to develop a logit recency bias, resulting in performance collapse. This aligns with the spirit of our proposed mechanism: the role of local layers is to produce a residual stream recency bias available to be picked up by the global NoPE logits; within the global NoPE logits, the model learns to pick up and maintain a consistent recency bias from the residual stream across heads to serve as its position encoding in attention.

\FloatBarrier\section{Discussion} \label{sec:discussion}
Despite their empirical success, how hybrid models interleaving SWA and global NoPE attention encode position in their global NoPE layers has remained a mystery. In this paper, we develop a novel mechanistic explanation for how such models accomplish this encoding. First, SWA creates a recency bias in its output by correlating neighboring embeddings, which emerges as a fundamental and in-built inductive bias present from initialization. Second, this recency bias is preserved by the intervening transformations (viz. residual stream addition, normalization, and MLPs). Third, it is selected by the global attention projections (to varying degrees across heads) and propagated to the global attention logits. Furthermore, this proposed mechanism may be extended to hybrid models that interleave linear layers like KDA in place of SWA (Appendix~\ref{app:kda}). This recency bias acts as a kind of \textit{implicit} relative position encoding, creating effects similar to explicit relative position encodings.

Our account of how such models encode position is particularly interesting given the special capacity of hybrid models with global NoPE attention for length extension and generalization. We hypothesize that this occurs for a few reasons. First, the window provides a finite-length local scale that does not dilute as sequence length increases, and which is invariant to the absolute sequence length. Also, the position encoding learned does not radically change at distances beyond those seen in training, as is the case with RoPE, implying that extrapolation to longer sequence lengths does not introduce any serious discontinuities. Finally, this position encoding may be uniquely flexible due to residing within the residual stream geometry, which may enable expressive and easily-learned interactions with content that may be adaptively read out by the global attention heads. 

Nevertheless, there are still many unknowns about how hybrid models with SWA and global NoPE attention encode position. For example, we still have not developed a complete quantitative account of how the recency bias propagates through all modules of the model, or how this affects the attenuation of the recency bias across depth and varying sequence lengths. Nor have we established a quantitative relationship between these characteristics and the extent of length generalizability. These topics are left to future work.

\bibliographystyle{style/tmlr}
\bibliography{references}

\clearpage
\onecolumn
\appendices
\startcontents[appendix]
{\begin{center}
    \Large Appendix
\end{center}}
{\hypersetup{linkcolor=black}\printcontents[appendix]{}{1}[2]{}}
\section{Measurement and Experimental Details}
\label{app:methods}

\subsection{Similarity measures and estimation}

We measure the cosine similarity after attention is added to the residual stream, except for isolated attention outputs at initialization and explicitly named module states. For the matrix plots, we average outer products of unit-length states before taking norms since this captures shared structure, though sampling noise can contribute. Our cosine and logit profiles use every integer lag from 1 to 4096.

\subsection{Models, training, and evaluation inputs}

\paragraph{SWA configurations.}
We study the 120M and 350M hybrid models described in Section~\ref{sec:thesis-empirics}, alternating SWA with global NoPE attention. SWA windows are $64,128,256,512,1024,2048,4096$; single-model comparisons use $w=128$. The SWA layers use RoPE unless stated otherwise. The two model scales were trained for 12B and 24B tokens, respectively, with a sequence length of 8192. Training and natural-language evaluation text use the GPT-2 tokenizer. The model embedding vocabulary is padded to 50,304 IDs. The figures in the main text use the 350M models and TextbookChapters unless their captions state otherwise.

\paragraph{KDA configurations.}
The KDA controls interleave KDA and global NoPE layers at both model scales. The supplementary KDA figures use final-checkpoint measurements on both TextbookChapters and random-token inputs; their aggregation conventions are stated below.

\paragraph{Evaluation support.}
Recency measurements use 48 sequences of length 8192. The natural-language inputs come from TextbookChapters; random-token controls sample IDs uniformly from the padded model vocabulary with a seed of 0. Queries are fixed at zero-based positions 4352--8191 for every lag $1\leq d\leq4096$, giving 184320 query--key pairs per lag. Initialization plots partition this support into four bins of 960 query positions. Validation-loss plots instead use DCLM sequences of length 4096 and show raw trajectories over the final 30\% of training.

\subsection{Aggregation and uncertainty conventions}
\label{app:aggregation-uncertainty}

\paragraph{Global NoPE logits.}
In the SWA and KDA profile and gap figures, we average the actual pre-softmax logits across heads at each lag. The logit gap is $\bar\ell(1)-\bar\ell(4096)$ from this head-mean profile. Where shown, shading spans the range of headwise gaps, not uncertainty.

\paragraph{Module changes and matrix measurements.}
Figure~\ref{fig:module-cosine-gap-floor} uses the final 350M RoPE-in-SWA, $w=128$ checkpoint. Each point is an after-minus-before change, averaged equally over the twelve SWA or twelve full-attention blocks. Residual addition steps compare with the isolated branch while block summaries compare the updated residual with its incoming residual. For Figure~\ref{fig:cross-moment-readout}, matrix norms are taken after averaging matrices across queries and sequences separately at each state and layer. Query--key pairs are not treated as independent samples.

\section{Supplement to Step 1: SWA Produces a Recency Bias in the Residual Stream} \label{app:step1}

This appendix provides additional theoretical and empirical results that corroborate those in Section~\ref{sec:step1}, using the same setting of 350M scale models using natural language inputs (TextbookChapters).

\subsection{Theory}

\paragraph{Mean-attention cross-moment derivation.}
\label{app:step1-theory}

As in Section~\ref{sec:step1-theory}, we assume that the norms of vectors entering and leaving the attention branch are concentrated, and therefore we work directly on unit-normalized moments $G_X$ and $G_Y$. We assume these moments are approximately stationary away from the causal boundary. For a fixed mean attention profile $a_d$, that is zero outside, $0\leq d<w$, let $y_i=A\sum_d a_d x_{i-d}$. Expanding both outputs gives
\[
\begin{aligned}
G_Y(d)
&=A\sum_{r,s=0}^{w-1}a_r a_s\,\mathbb{E}[x_{i-r}x_{i-d-s}^{\top}]A^\top\\
&=A\sum_{r,s=0}^{w-1}a_r a_s\,G_X(d+s-r)A^\top\\
&=A\sum_{k=-(w-1)}^{w-1}g_a(k)G_X(d-k)A^\top,
\qquad g_a(k)=\sum_s a_s a_{s+k}.
\end{aligned}
\]
Here $d$ is the output lag and $k=r-s$ is the difference between two filter offsets. For learned attention, taking $a_r=\mathbb{E}[\alpha_i(r)]$ is a fixed-profile approximation that omits weight fluctuations and their dependence on the inputs.

For uncorrelated white inputs, $G_X(d)=\Sigma\mathbbm{1}\{d=0\}$, and therefore $G_Y(d)=g_a(d)A\Sigma A^\top$. If we further assume attention is uniform, then the triangular profile introduced in the text is a special case of the same convolution.

\subsection{Empirics}

\begin{figure}[!htbp]
\centering
\includegraphics[width=\textwidth]{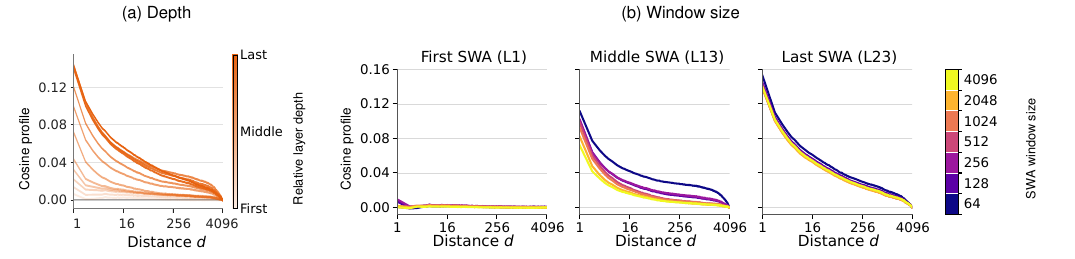}
\caption{\textit{Residual cosine at 350M.} Post-attention gap $c(d)-c(4096)$: (a) every SWA layer at the final $w=128$ checkpoint; (b) first, middle, and last SWA layers across windows 64--4096.}
\label{fig:residual-depth-windows-350m}
\end{figure}

\paragraph{Depth-wise and window-size dependence.} \label{app:swa-depth-windows}
As noted in Section~\ref{sec:step1-empirics}, the residual stream recency bias accumulates across depth. To investigate this at finer resolution, Figure~\ref{fig:residual-depth-windows-350m}a shows the residual stream recency profile following every SWA layer at the last checkpoint of training, which demonstrates that the trend of increasing recency bias through depth is notably smooth across layers. Comparing Figure~\ref{fig:residual-depth-windows-350m}a to its 120M-scale companion Figure~\ref{fig:120m-robustness-loss-composite}c in Appendix~\ref{app:120m}, it is particularly notable how consistent this layer-wise recency bias trend is across model scales, not only in ordering, but also in magnitude. This supports that the residual stream recency bias provides a functional role regardless of scale.

To supplement the comparison of window sizes in Figure~\ref{fig:residual-window-gaps}, Figure~\ref{fig:residual-depth-windows-350m}b displays the full cosine profiles at the first, middle, and last SWA layers of different window sizes; Figure~\ref{fig:swa-loss-windows-350m} in the main text shows the validation loss curves for the same range of window sizes. Together, these results show that varying window size has a clear and predictable effect on both cosine profiles and language modeling (though we cannot fully isolate its causal role).

\FloatBarrier\section{Supplement to Step 3: Global NoPE Inherits the Residual Stream Recency Bias} \label{app:step3}

This appendix examines how the recency structure induced by SWA and shaped by the intervening transformations analyzed in Section~\ref{sec:step2} is read out by global NoPE. It provides a cross-moment analysis and additional measurements supporting the readout in Section~\ref{sec:step3}. We also explain how cross-moment magnitude and alignment help us understand the recency bias in both the residual stream and logits.

\subsection{Theory}

\paragraph{Decomposing the general similarity structure.}
To analyze this readout, we study the full lagged cross-moment. Central to our understanding of this general structure is that different slices of it reduce to the cosine similarity and per-head logit curves, via the Frobenius inner product. Just as a vector inner product depends on both magnitude and alignment, it is instructive to decompose the Frobenius inner product into the contributions of cross-moment magnitude and its vectorized alignment with either the isotropic component $I$ or the composed query-key projection $M=W_Q^\top W_K/\sqrt{D_H}$.

For nonzero matrices $G$ and $X$ of the same shape, the Frobenius inner product decomposes as
\[
    \langle G,X\rangle_F
    = \|G\|_F\|X\|_F\cos(G,X),
    \qquad
    \cos(G,X)=\frac{\operatorname{tr}(G^\top X)}{\|G\|_F\|X\|_F}.
\]
Hence, the matrix cosine is just the ordinary cosine between the vectorized matrices: the two norms set the scale, while their alignment determines the sign.

To see how this pertains to a recency bias, let $B_X=G_X(1)-G_X(L)$, and $B_Z=G_Z(1)-G_Z(L)$ for the actual post-normalization input $Z$ to global attention. Taking $X=I$ or $X=M$ gives
\[
\begin{aligned}
    c(1)-c(L) &= \sqrt{D}\,\|B_X\|_F\cos(B_X,I),\\
    \ell(1)-\ell(L) &= \|B_Z\|_F\|M\|_F\cos(B_Z,M).
\end{aligned}
\]
Here $\|I\|_F=\sqrt{D}$, and $\|M\|_F$ is the head's readout gain. Therefore, the recency gaps for both cosine similarity and the global NoPE logits depend on the underlying matrix lag contrast of the cross-moment, which is unsigned (i.e., a large magnitude alone does not imply a positive cosine or logit gap). Meanwhile, its vectorized cosine alignment determines the sign. Next, we empirically assess its magnitude and alignment.

\subsection{Empirics} \label{app:step3-empirics}

\begin{figure}[!htbp]
\centering
\includegraphics[width=0.80\textwidth]{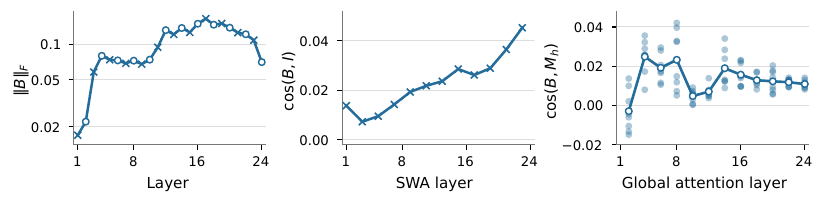}
\caption{\textit{Cross-moment magnitude and readout alignment.} Final 350M RoPE-in-SWA model ($w=128$). Left: $\|B_X\|_F$ at post-block residual states; crosses mark SWA and circles global NoPE layers. Middle: $\cos(B_X,I)$ at SWA layers. Right: $\cos(B_Z,M)$ at model-scale post-LN1 inputs to global heads; small points show eight heads and connected circles their unweighted mean. Magnitude uses a logarithmic axis; alignments use linear axes.}
\label{fig:cross-moment-readout}
\end{figure}

\paragraph{Cross-moment magnitude and readout alignment.}
For a broader view on the cross-moment magnitude and its alignment to both the isotropic component and per-head projections, Figure~\ref{fig:cross-moment-readout} measures the matrix lag contrast $B$ of the cross-moment and its cosine similarity with $I$ and $M$ across depth. Indeed, as is expected from the separate analyses of residual stream cosine and the logits, the total similarity magnitude increases through depth along with its alignment to $I$ (which together imply the increased recency bias through depth we see in Section~\ref{sec:step1}); meanwhile, the global NoPE projections maintain a positive alignment with $B$ that remains relatively constant, producing positive and consistent logit gaps across depth (as seen in Section~\ref{sec:step3-empirics}).

\paragraph{Depth-wise and window-size dependence.}
\label{app:logit-depth-windows}

\begin{figure}[!htbp]
\centering
\includegraphics[width=\textwidth]{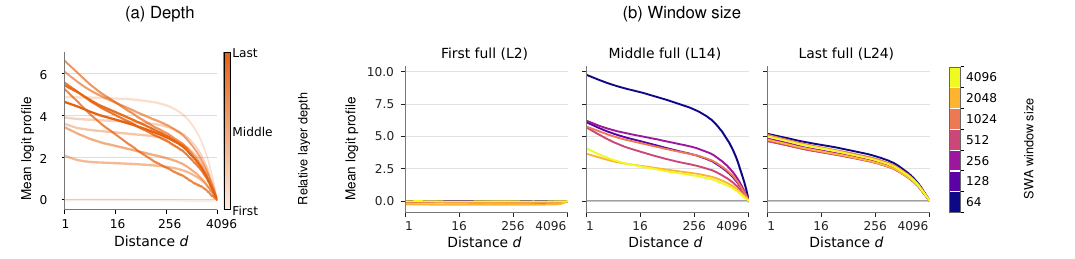}
\caption{\textit{Global NoPE logits at 350M.} Head-mean gap $\bar\ell(d)-\bar\ell(4096)$: (a) every global layer at the final $w=128$ checkpoint; (b) first, middle, and last global layers across window sizes.}
\label{fig:logit-depth-windows-350m}
\end{figure}

For a more fine-grained view of depth-wise patterns than provided in Figure~\ref{fig:logit-gap-windows}, Figure~\ref{fig:logit-depth-windows-350m}a shows the mean logit profile at every global NoPE layer. As noted in Section~\ref{sec:step3-empirics}, the pattern is different from that of the residual stream recency bias: after an initial rise, the logit recency bias is largely consistent across layers. Similarly, comparing the logit profiles for different window sizes, Figure~\ref{fig:logit-depth-windows-350m}b shows a weaker relationship between recency bias and window size than in the residual stream. The relative lack of variation in the logit profiles with depth and window size suggests that, as we discussed in Section~\ref{sec:step3-empirics}, the model seeks to develop a consistent logit-based recency bias to serve as its position encoding in attention.

\FloatBarrier

\section{Controls}\label{app:controls}

\subsection{120M-scale robustness} \label{app:120m}

\begin{figure}[p]
\centering
\includegraphics[page=1,width=\textwidth]{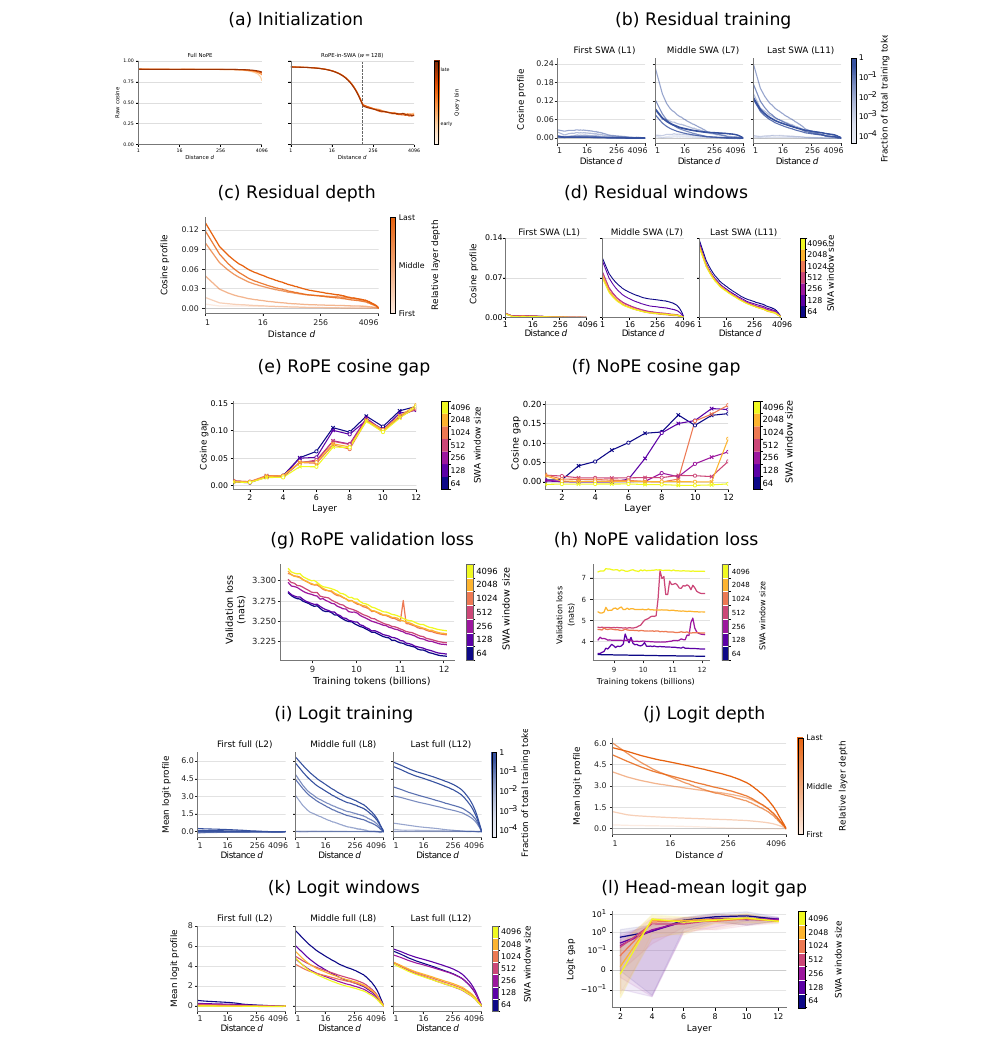}
\caption{\textit{120M TextbookChapters robustness and validation loss.} (a) Attention-branch cosine at initialization. (b--f) Post-attention residual cosine during training, across depth and SWA windows, including RoPE- and NoPE-in-SWA gaps. (g--h) Validation loss over the final 30\% of training on 4096-token sequences. (i--l) Global-NoPE head-mean logit profiles and gaps during training, across depth and windows. Profile gaps are relative to lag 4096; window sizes are 64--4096. Window comparisons do not isolate the contribution of recency to the loss.}
\label{fig:120m-robustness-loss-composite}
\end{figure}

To verify that the 350M-scale results of the main text are robust across model scales, we train 120M models with the same configurations of SWA window size and varying RoPE and NoPE within the window, on the same dataset and for a number of tokens comparable given the scale difference  (see Appendix~\ref{app:methods}). These 120M-scale results corroborate the main-text 350M-scale results, with the same qualitative conclusions from each panel provided in the composite of Figure~\ref{fig:120m-robustness-loss-composite}. That is, SWA's recency bias is present from initialization (a), strengthens with training and depth (b--c), becomes sharper with smaller windows (d--e), and propagates to and is selected by the logits (i--l).

Furthermore, Figure~\ref{fig:120m-robustness-loss-composite}f--h displays the cosine gap for NoPE-in-SWA, as well as a comparison of the validation losses for both RoPE and NoPE within the window. Of particular note is that, in both cases, loss decreases with smaller windows despite fewer training flops, indicating that smaller windows may be beneficial, perhaps due to their ability to resolve a strong recency bias. In further support of this, with NoPE-in-SWA loss collapses entirely for larger windows; this is suggestive that, without a sufficiently strong architectural mechanism to resolve a recency bias and therefore encode position, such models fail to learn language effectively.

\subsection{Random-token controls} \label{app:random-controls}

\begin{figure}[p]
\centering
\includegraphics[page=2,width=\textwidth]{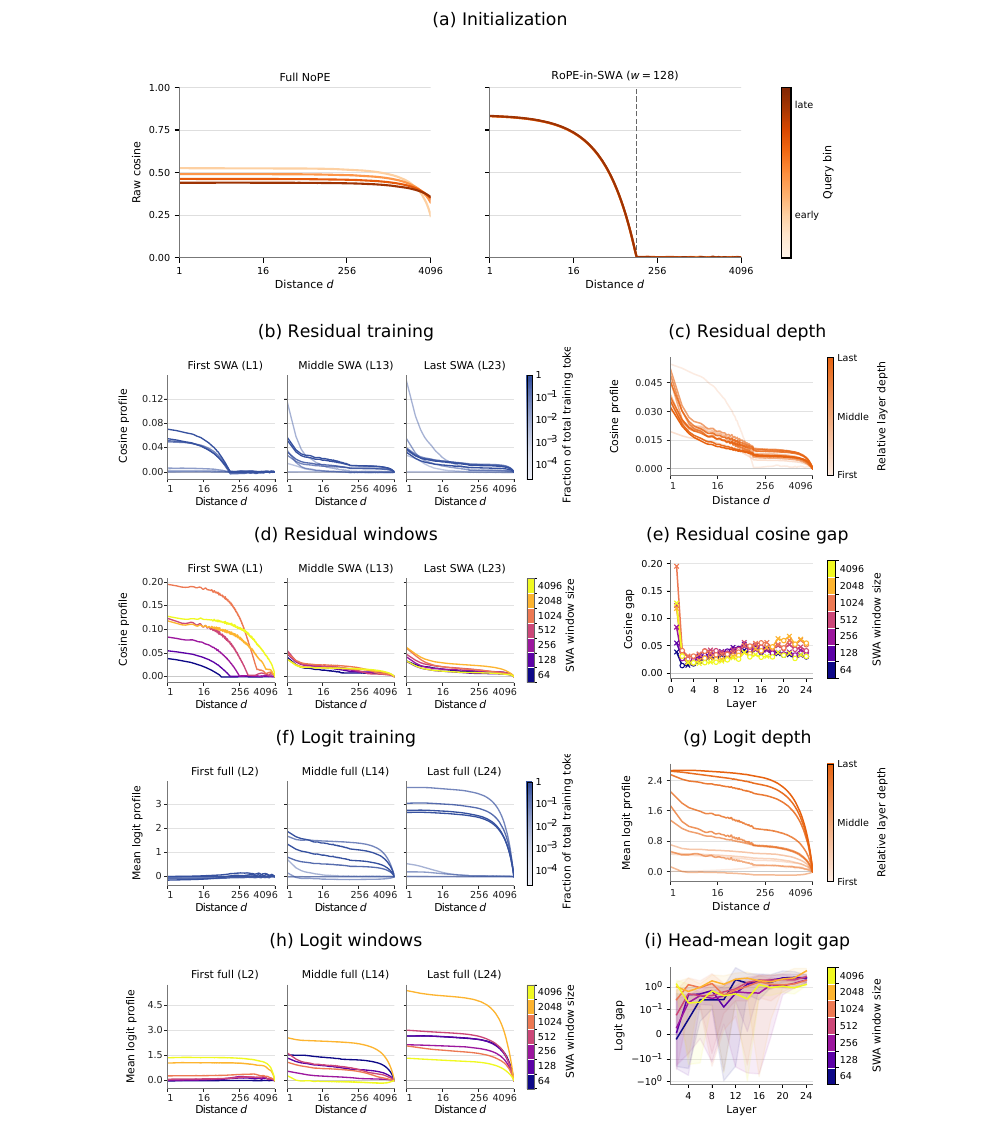}
\caption{\textit{Random-token control: 350M, RoPE within SWA.} (a) Attention-branch cosine at initialization. (b--e) Post-attention residual cosine during training, across depth and windows, including the depthwise gap. (f--i) Global-NoPE head-mean logit profiles and gaps during training, across depth and windows. Profile gaps are relative to lag 4096; window sizes are 64--4096.}
\label{fig:random-350m-rope-composite}
\end{figure}

\begin{figure}[p]
\centering
\includegraphics[page=3,width=\textwidth]{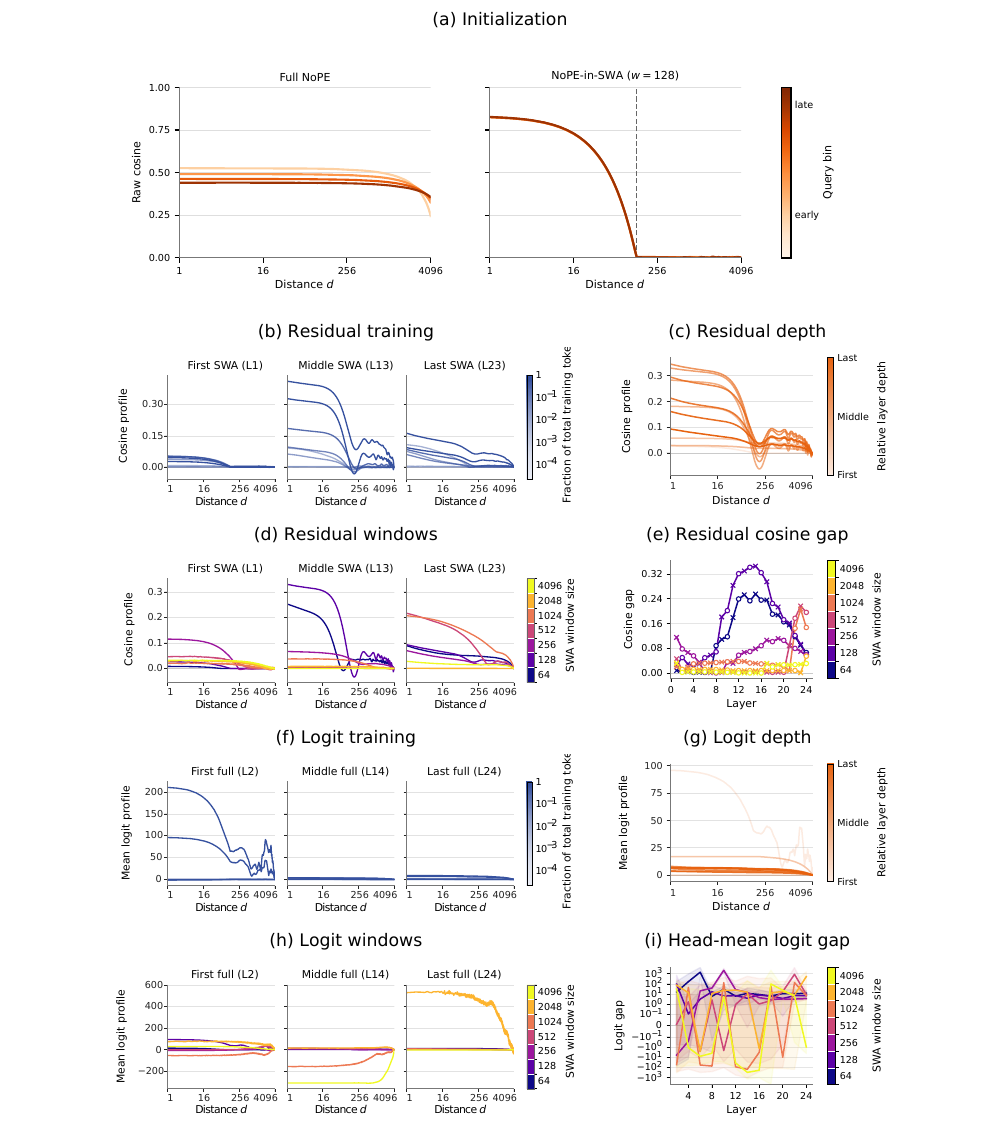}
\caption{\textit{Random-token control: 350M, NoPE within SWA.} (a) Attention-branch cosine at initialization. (b--e) Post-attention residual cosine during training, across depth and windows, including the depthwise gap. (f--i) Global-NoPE head-mean logit profiles and gaps during training, across depth and windows. Profile gaps are relative to lag 4096; window sizes are 64--4096.}
\label{fig:random-350m-nope-composite}
\end{figure}

\begin{figure}[p]
\centering
\includegraphics[page=4,width=\textwidth]{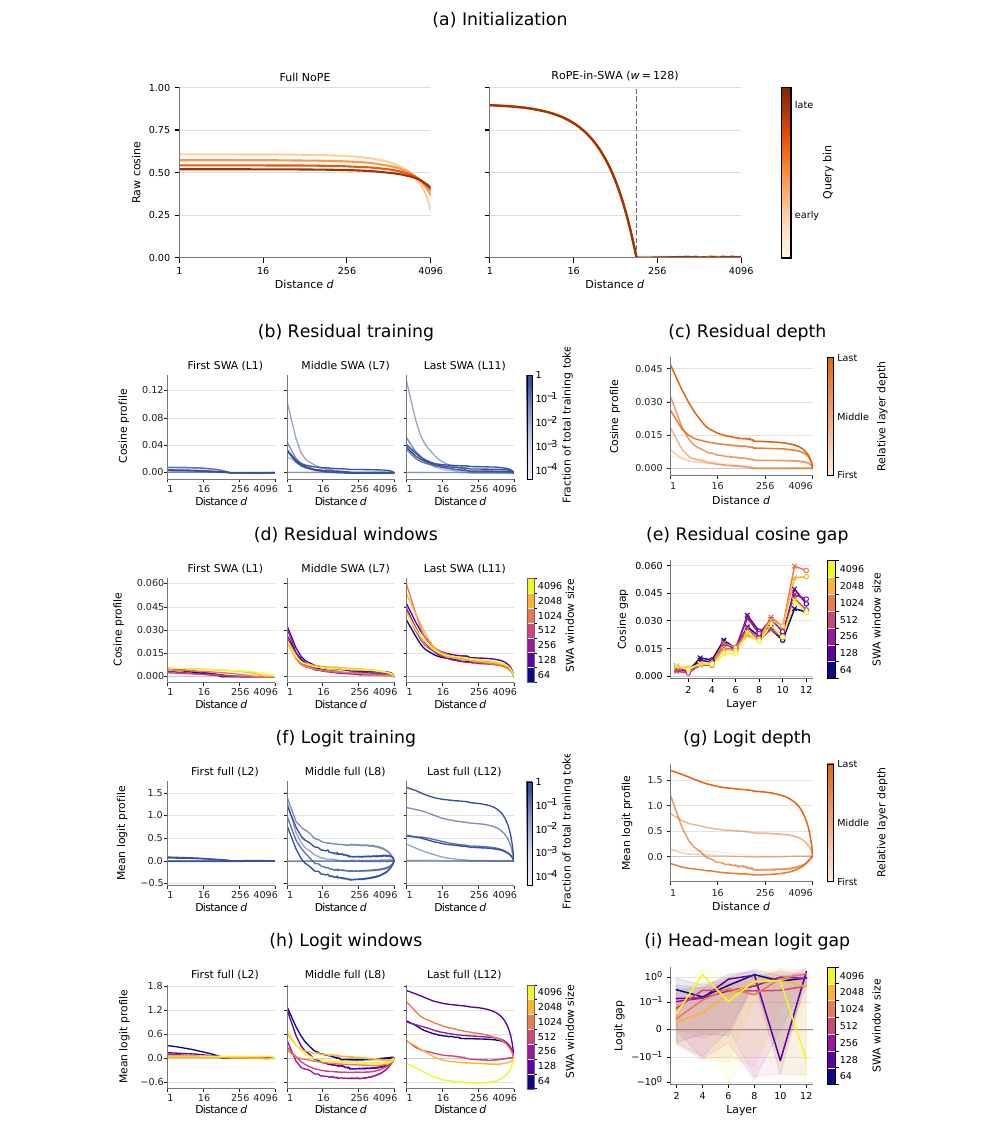}
\caption{\textit{Random-token control: 120M, RoPE within SWA.} (a) Attention-branch cosine at initialization. (b--e) Post-attention residual cosine during training, across depth and windows, including the depthwise gap. (f--i) Global-NoPE head-mean logit profiles and gaps during training, across depth and windows. Profile gaps are relative to lag 4096; window sizes are 64--4096.}
\label{fig:random-120m-rope-composite}
\end{figure}

\begin{figure}[p]
\centering
\includegraphics[page=5,width=\textwidth]{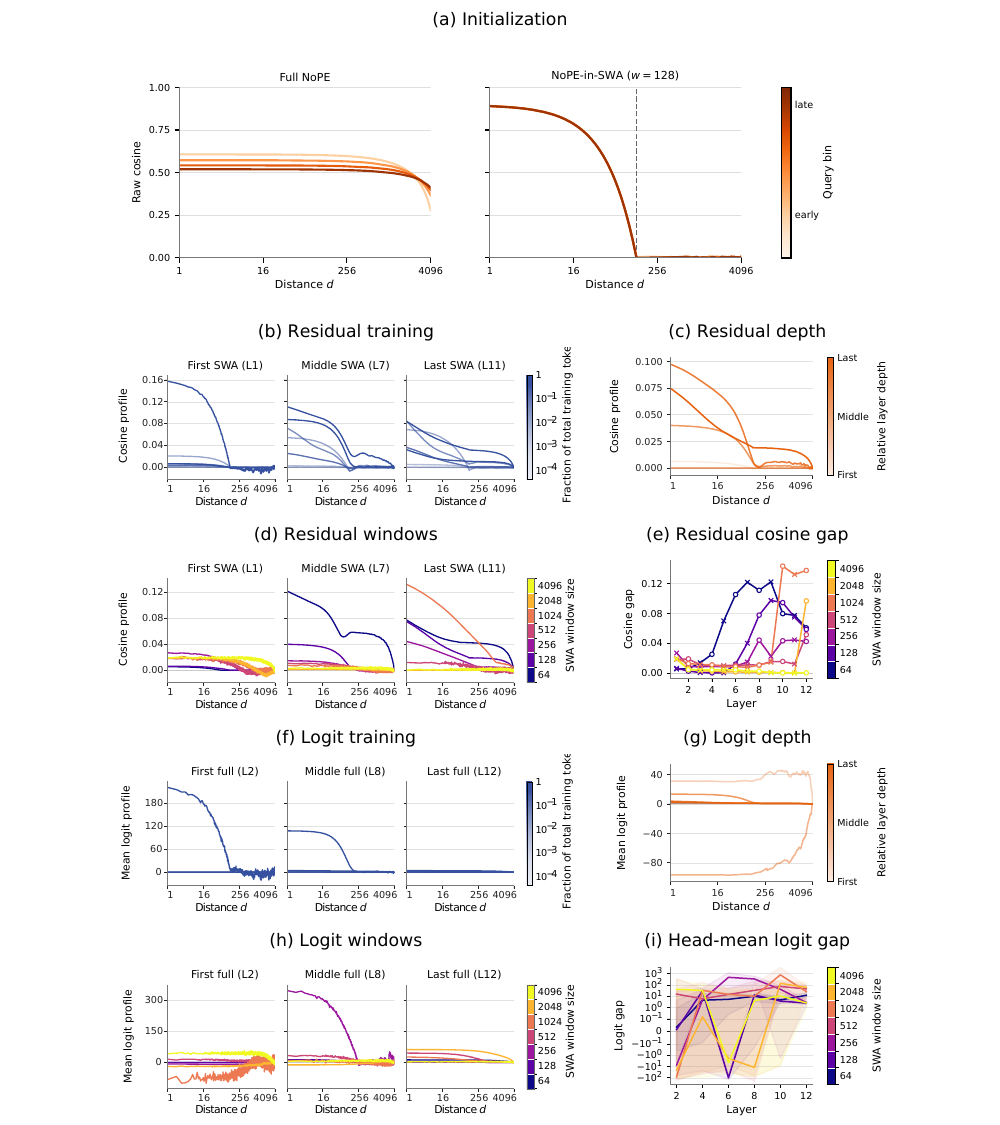}
\caption{\textit{Random-token control: 120M, NoPE within SWA.} (a) Attention-branch cosine at initialization. (b--e) Post-attention residual cosine during training, across depth and windows, including the depthwise gap. (f--i) Global-NoPE head-mean logit profiles and gaps during training, across depth and windows. Profile gaps are relative to lag 4096; window sizes are 64--4096.}
\label{fig:random-120m-nope-composite}
\end{figure}

The main analyses use natural-language sequences, as these are the natural domain of the language models we train. However, language has an underlying correlation structure that may confound the evidence for our proposed architectural mechanism. Hence, repeating the complete SWA analysis on uniformly sampled tokens from the model's vocabulary is an important control to isolate the structural contribution from natural-language statistics. The following controls use the same query positions and lags as the TextbookChapters analyses and report results for both RoPE and NoPE within SWA (Figures~\ref{fig:random-350m-rope-composite}--\ref{fig:random-120m-nope-composite}; each composite groups the observables for one SWA encoding and model scale).

These controls also corroborate the main-text findings. Across all four composites, we observe the recency bias present at initialization and strengthening with training and depth (a--c). SWA's position encoding  has a larger impact on the effect of window size (d--e): for NoPE-in-SWA at both model scales, we observe the same pattern of smaller windows promoting stronger recency biases; however, for RoPE-in-SWA at both scales, the trends were inconclusive, likely due to random tokens altering the learned RoPE decay profile. Finally, across all four settings, the recency bias transfers to the logits particularly for smaller windows, but this transfer is less consistent than in natural language (f--i). Overall, these results are consistent with our proposed architectural mechanism functioning, but being untuned for data outside its training distribution.  

\FloatBarrier

\section{KDA extension} \label{app:kda}

\subsection{Theory}

KDA updates its recurrent state as
\[
S_t=(I-\beta_t k_tk_t^\top)\operatorname{Diag}(\alpha_t)S_{t-1}+\beta_t k_tv_t^\top.
\] Here, $\alpha_t$ controls how much of each state channel is retained, while the delta update revises the association with the current key. A write from position $i$ reaches position $t$ through the intervening transition products. We now use a simplified approximation to show how this repeated mixing can produce a recency bias, following the same intuition as for SWA.

\paragraph{Intuition.}
Consider a simplified scenario in which the projected values $o_t$ are zero-mean, of similar magnitude, and uncorrelated across positions, and the effective contribution of a past value decreases by a fixed factor $0<\lambda<1$ at each step. Averaging over the content-dependent gates, writes, and query readout, we approximate the attention output before normalization and gating as an exponential moving average,
\[
y_t \approx \lambda y_{t-1}+(1-\lambda)o_t
= (1-\lambda)\sum_{r=0}^{\infty}\lambda^r o_{t-r},
\]
where the infinite sum describes positions sufficiently far from the start of the sequence. As in SWA, nearby outputs are similar because they combine many of the same values; consider, for example, $y_t$ and $y_{t-d}$. In particular, all values contributing to $y_{t-d}$ also contribute to $y_t$, but with their weights reduced by $\lambda^d$. Hence, if output norms are concentrated, their expected cosine similarity is approximately
\[
c_{\mathrm{KDA}}(d) \approx \lambda^d.
\]
Thus, the same shared-values argument that gives a triangular profile for uniform SWA gives an exponential profile in this simplified KDA case.

\paragraph{Length scaling.}
The retention factor plays a role analogous to the SWA window size: a smaller $\lambda$ produces a narrower, steeper recency profile, while a larger $\lambda$ retains correlations over longer distances. Critically, like SWA, this scale is length-independent and therefore remains equally well resolved with sequence length.

\paragraph{Correlated inputs.}
As in the general SWA case, the incoming similarity structure also affects the output. Let $G_O(d)$ and $G_Y(d)$ denote the lagged cross-moments of the unit-normalized projected values and outputs, respectively. Assume norms are concentrated, then for the same exponential weights $a_r=(1-\lambda)\lambda^r$, the filter calculation then gives the approximate relation
\[
G_Y(d)\propto\sum_{k\in\mathbb{Z}}g_a(k)G_O(d-k),
\qquad g_a(k)=\frac{1-\lambda}{1+\lambda}\lambda^{|k|}.
\]
Thus, the incoming structure is mixed across distances by a smoothly decaying overlap profile. As in SWA, the subsequent transformations determine how this structure enters the residual stream and whether later NoPE projections read it out as larger logits for recent keys.

\subsection{Empirics}

We empirically test whether our proposed mechanism extends to KDA-interleaved models at different scales, when using both natural language data and random-token controls. Figures~\ref{fig:kda-textbook-composite} and~\ref{fig:kda-random-composite} below show the KDA results; each row compares both model scales, and all logit panels use the head-mean convention specified in Appendix~\ref{app:aggregation-uncertainty}. Indeed, the same empirical pattern appears: KDA layers primarily increase residual stream recency bias across depth (a--d); and later NoPE layers learn recency-biased logits (e--h). This indicates that the general mechanism we identify is not specific to SWA, but can apply to many local sequence mixers including the variety of increasingly common linear gated attention types.

\begin{figure}[p]
\centering
\includegraphics[page=6,width=\textwidth]{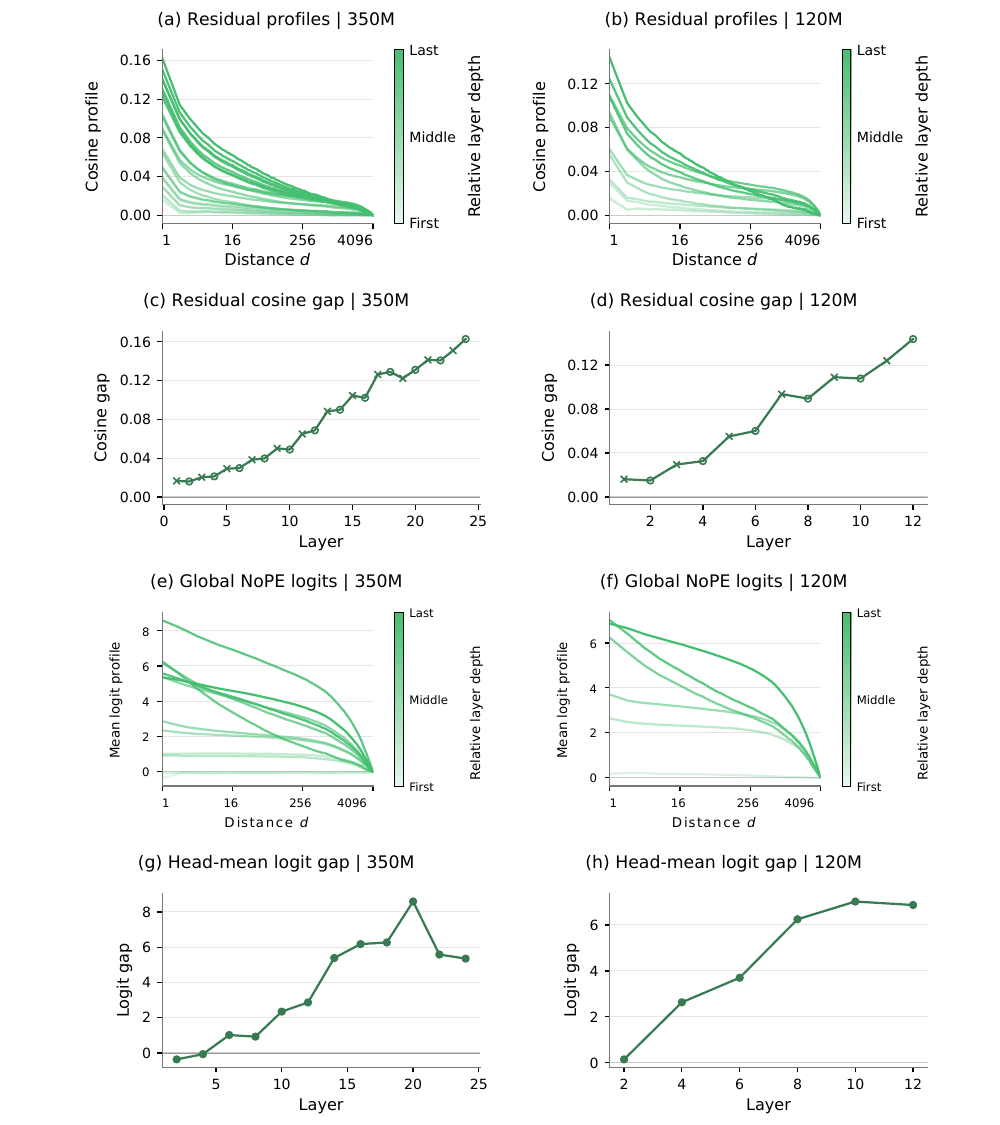}
\caption{\textit{KDA--NoPE natural-language extension.} Columns compare 350M (left) and 120M (right). (a--b) Residual cosine gap profiles across depth; (c--d) lag-1 residual gaps; (e--f) global-NoPE head-mean logit gap profiles; (g--h) lag-1 head-mean logit gaps. All gaps are relative to lag 4096.}
\label{fig:kda-textbook-composite}
\end{figure}

\begin{figure}[p]
\centering
\includegraphics[page=7,width=\textwidth]{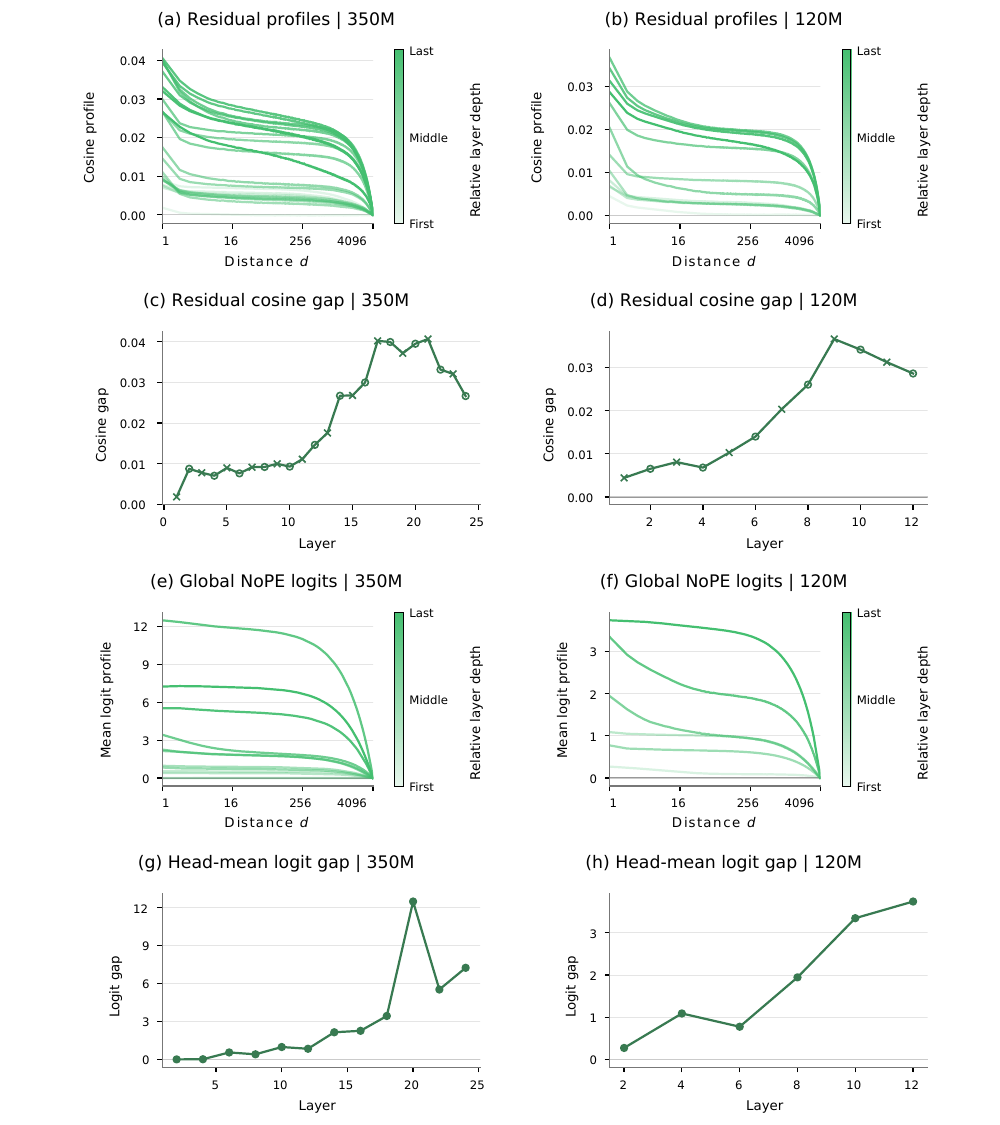}
\caption{\textit{KDA--NoPE random-token extension.} Columns compare 350M (left) and 120M (right). (a--b) Residual cosine gap profiles across depth; (c--d) lag-1 residual gaps; (e--f) global-NoPE head-mean logit gap profiles; (g--h) lag-1 head-mean logit gaps. All gaps are relative to lag 4096.}
\label{fig:kda-random-composite}
\end{figure}

\end{document}